\documentclass{article}
\usepackage{arxiv}

\usepackage[utf8]{inputenc} 
\usepackage[T1]{fontenc}    
\usepackage{hyperref}       
\usepackage{url}            
\usepackage{booktabs}       
\usepackage{amsfonts}       
\usepackage{amsmath,amssymb}
\usepackage{nicefrac}       
\usepackage{microtype}      
\usepackage{graphicx}
\usepackage{enumitem}
\usepackage{bm}
\usepackage[table]{xcolor}
\usepackage{siunitx}
\usepackage{caption}
\usepackage{array}
\usepackage{float}
\definecolor{headpink}{RGB}{255,245,250}  

\definecolor{headgray}{gray}{0.95}
\definecolor{extrow}{gray}{0.96}
\definecolor{mypink}{RGB}{255, 105, 180} 
\definecolor{myblue}{RGB}{30, 144, 255}

\usepackage{expl3}
\usepackage{xparse}

\ExplSyntaxOn

\int_new:N \l_my_ratio_int 
\int_new:N \l_my_idx_int
\int_new:N \l_my_len_int

\NewDocumentCommand{\discretegradient}{m}{
    
    \tl_set:Nn \l_tmpa_tl { #1 }
    
    \int_set:Nn \l_my_len_int { \tl_count:N \l_tmpa_tl }
    
    \int_zero:N \l_my_idx_int

    \tl_map_inline:Nn \l_tmpa_tl {
        \int_compare:nNnTF { \l_my_len_int } > { 1 }
        {
            \int_set:Nn \l_my_ratio_int { 100 - ( \l_my_idx_int * 100 / (\l_my_len_int - 1) ) }
        }
        {
            \int_set:Nn \l_my_ratio_int { 100 } 
        }
        
        \textcolor{mypink!\int_use:N \l_my_ratio_int!myblue}{##1}
        
        \int_incr:N \l_my_idx_int
    }
}
\ExplSyntaxOff

\title{\texorpdfstring{\discretegradient{Neuro-Vesicles}}{Neuro-Vesicles}: Neuromodulation Should Be a Dynamical System, Not a Tensor Decoration}

\author{
  Zilin Li\thanks{First author.}\\
  Donghua University\\
  2999 North Renmin Road 201620 \\
  Shanghai, China\\
  \texttt{tzulamlee@gmail.com} %
  \AND   %
  Weiwei Xu \\
  Donghua University \\
  2999 North Renmin Road 201620 \\
  Shanghai, China\\
  \texttt{yumuzaijie@gmail.com} \\
  \And   %
  Vicki Kane \\
  Park University \\
  8700 NW River Park Dr. \\
  Parkville, MO 64152, USA \\
  \texttt{1745807@park.edu} \\
}

\begin{document}
\maketitle

\begin{abstract}
We introduce \emph{Neuro-Vesicles}, a formal framework that augments conventional neural networks with a previously missing computational layer: a dynamical population of mobile, discrete entities that live \emph{alongside} the network rather than \emph{inside} its tensors. Unlike classical neuromodulation mechanisms—which collapse modulation into scaling factors, gating masks, or low-rank parameter shifts—Neuro-Vesicles treat modulation as an explicit stochastic process unfolding on the network graph. Each vesicle is a self-contained object carrying a vector-valued payload, a type label, an internal state, and a finite lifetime. Vesicles are emitted from nodes in response to activations, errors, or meta-signals; migrate across the graph according to learned transition kernels; probabilistically dock at individual nodes; locally alter activations, parameters, learning rules, or external memory through content-dependent release operators; and finally decay or are absorbed.

This event-based interaction layer fundamentally reshapes modulation: vesicles can accumulate, disperse, trigger cascades, carve transient computational pathways, or write structured traces into topological memory. Rather than applying a fixed modulation function at every forward pass, the system exhibits multi-timescale behavior where fast neural computation is continuously shaped by slowly evolving vesicle dynamics. In the limit of many vesicles with short lifetimes, the framework recovers familiar tensor-level conditioning layers such as FiLM, hypernetworks, or attention; in the opposite limit of sparse, long-lived vesicles, it approaches a discrete symbolic system of mobile “agents’’ that edit computation only at rare but decisive moments.

We present a complete mathematical specification of the Neuro-Vesicle paradigm, including the base network as a directed computational graph; the vesicle state space; emission, migration, docking, release, and decay kernels; the coupled stochastic dynamics linking vesicles and network behavior; a continuous density relaxation for differentiable training; and a reinforcement-learning formulation that treats vesicle control as an overlay policy. We then outline how the same formalism naturally extends to spiking neural networks and neuromorphic hardware such as the Darwin3 chip, yielding a three-part computational architecture—a parameterized network, a topological external memory, and a mobile event-driven dynamical layer—that turns neuromodulation from a tensor decoration into a first-class mechanism of computation.
\end{abstract}

\section{Base Network as a Graph}

\begin{figure}[H]
    \centering
    \includegraphics[width=\linewidth]{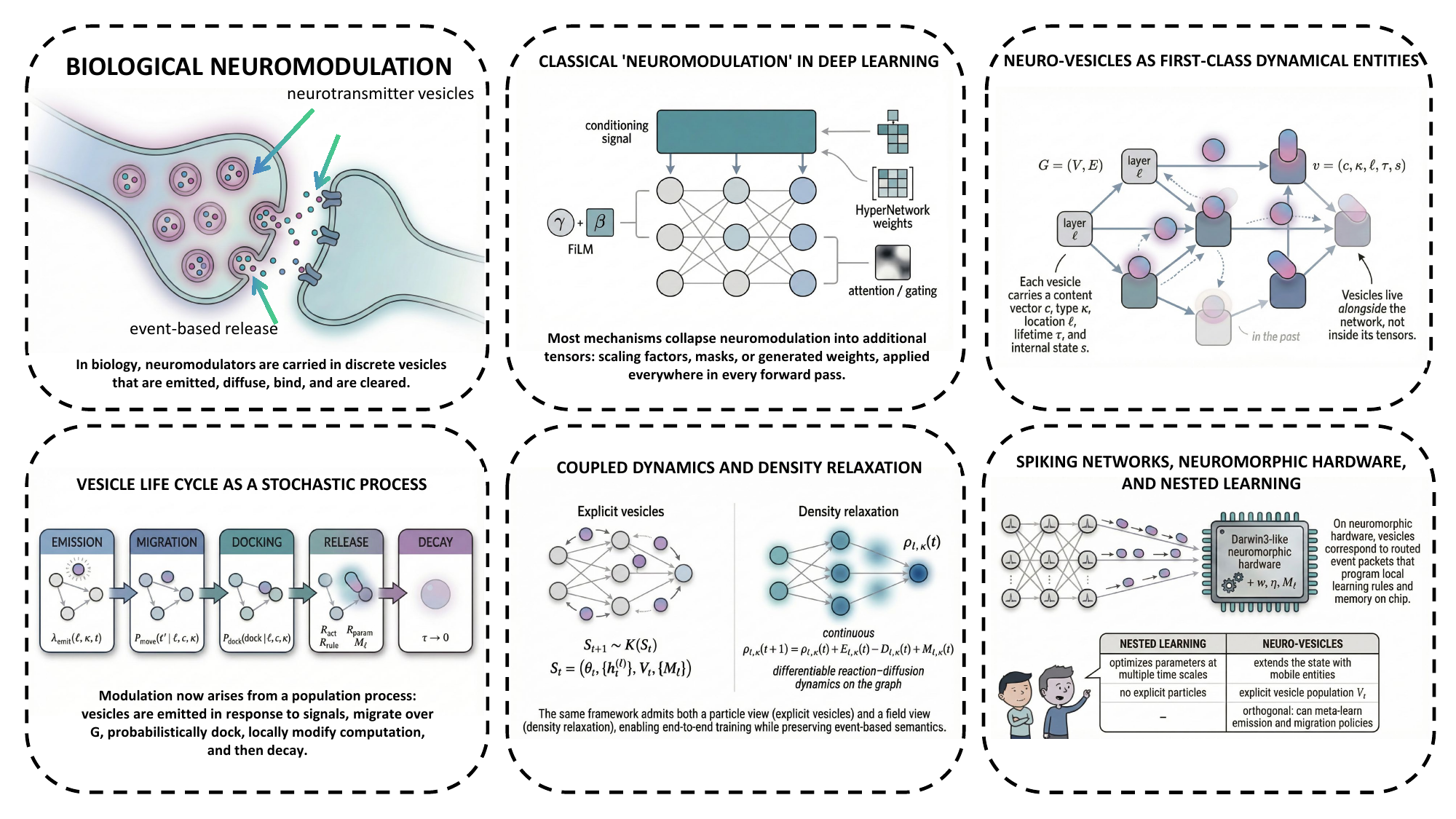} 
    \caption{%
    Overview of the Neuro-Vesicle (NV) paradigm.
    From left to right, the panels illustrate:
    \textbf{(a)} biological neuromodulation via neurotransmitter vesicles at chemical synapses;
    \textbf{(b)} classical ``neuromodulation'' in deep learning implemented as additional tensors
    (conditioning signals, FiLM layers, HyperNetwork-generated weights, and attention/gating);
    \textbf{(c)} NVs as first-class dynamical entities---mobile vesicles
    \(v=(\mathbf{c},\kappa,\ell,\tau,s)\) that move on the network graph \(G=(V,E)\);
    \textbf{(d)} the vesicle life cycle as a stochastic process
    (emission, migration, docking, release, decay);
    \textbf{(e)} coupled dynamics of explicit vesicles and their continuous
    density relaxation on the graph;
    \textbf{(f)} deployment in spiking neural networks and neuromorphic hardware,
    and the orthogonal relationship between NVs and nested learning.
    The bottom legend summarizes the core notation:
    \emph{Neuro-Vesicle} \(=\)(content \(\mathbf{c}\), type \(\kappa\), location \(\ell\),
    lifetime \(\tau\), state \(s\)), \(G=(V,E)\) is the network graph substrate,
    and modulation is emphasized as emerging from vesicle dynamics rather than
    from extra tensors alone.
    }
    \label{fig:nv_overview}
\end{figure}

Figure~\ref{fig:nv_overview} provides a high-level overview of the
Neuro-Vesicle paradigm, from biological neuromodulation to deployment
on spiking and neuromorphic systems.

The starting point of our framework is deliberately conservative: we do
not alter the structure of a standard neural network.
Instead, we \emph{re-interpret} it as a substrate on which an additional
dynamical layer can live.
This avoids confusing the proposed mechanism with yet another architectural
tweak, and makes it clear that Neuro-Vesicles are not a replacement for
existing models, but an overlay that can, in principle, be added to
almost any architecture.

Let a standard feedforward or recurrent neural network be written as a
function
\begin{equation}
    f_\theta : \mathcal{X} \to \mathcal{Y},
\end{equation}
with parameters \(\theta\) and training distribution
\((x,y) \sim \mathcal{D}\).
We view the network as a directed graph
\begin{equation}
    G = (V,E),
\end{equation}
where each node \(v \in V\) corresponds to a computational module
(e.g., a single neuron, a channel, a layer, or a transformer block) and
each edge \((u,v) \in E\) denotes information flow.
This graph view is intentionally generic: it covers convolutional stacks,
transformers, mixture-of-experts, and even more irregular architectures
such as neural program interpreters.

For notational clarity, we assume the network can be decomposed into
\(L\) ordered layers
\begin{equation}
    h^{(0)} = x,\qquad
    h^{(l)} = \phi^{(l)}\!\big(h^{(l-1)}; \theta^{(l)}\big),
    \quad l=1,\dots,L,
\end{equation}
with output \(f_\theta(x) = h^{(L)}\).
Here \(h^{(l)}\) is the hidden representation at layer \(l\),
\(\theta^{(l)}\) are parameters local to that layer, and
\(\phi^{(l)}\) is a deterministic mapping (e.g., affine transformation
plus nonlinearity, attention block, etc.).
The details of \(\phi^{(l)}\) are deliberately abstracted away, because
Neuro-Vesicles are defined to act on this layer-wise structure without
constraining its internal implementation.

We write the standard training objective as
\begin{equation}
    L_{\text{base}}(\theta)
    = \mathbb{E}_{(x,y) \sim \mathcal{D}}
      \big[\,\ell\big(f_\theta(x), y\big)\,\big],
\end{equation}
where \(\ell(\cdot,\cdot)\) is a suitable loss function.
In almost all existing work on ``neuromodulation'' in deep learning,
the effect of modulatory signals is implemented as additional terms inside
this forward mapping: scaling factors on activations, additive biases,
hypernetworks that generate weights, context-dependent attention masks,
and so on.\footnote{See, e.g., FiLM layers~\cite{perez2018film} and
HyperNetworks~\cite{ha2016hypernetworks} for two influential families of
such mechanisms.}
Conceptually, all of these live \emph{inside} the tensors and weight
matrices, and are updated at the same time scale and with the same
gradient flow as the rest of the network.

By contrast, Neuro-Vesicles are designed to form a \emph{separate}
dynamical system that runs on top of the graph \(G\), interacting with
but not being reduced to the usual tensor algebra.
The aim is to more faithfully capture the idea that neuromodulation is
an event-driven, spatially localized, and temporally extended process,
rather than just a few extra channels in the forward pass.

\section{State Space of Neuro-Vesicles}

\subsection{Single Vesicle State}

We now describe the basic unit of the proposed dynamical layer.
A \emph{Neuro-Vesicle} is a discrete entity
\begin{equation}
    v = (\mathbf{c}, \kappa, \ell, \tau, s),
    \label{eq:vesicle_state}
\end{equation}
with the following components:
\begin{itemize}[leftmargin=2em]
    \item \(\mathbf{c} \in \mathbb{R}^{d_c}\):
        vesicle content vector (payload), conceptually an artificial
        neuromodulator embedding.
        This can be thought of as the analogue of the ``chemical content''
        of a biological vesicle, but here it is entirely abstract and
        learnable.
    \item \(\kappa \in \{1,\dots,K\}\):
        vesicle type (e.g., ``families'' analogous to different
        biological neuromodulators).
        Different types may use different emission rules, migration
        policies, or release operators.
    \item \(\ell \in V\):
        current location on the network graph (node, layer, or module).
        This makes vesicles spatially grounded: they do not act
        everywhere at once, but only at the nodes where they happen
        to be.
    \item \(\tau \in \mathbb{R}^+\):
        remaining lifetime.
        The lifetime induces a natural temporal extent for the influence
        of a vesicle; once it expires, its direct effects stop, although
        its past interactions may have long-term consequences via learned
        parameters and memory.
    \item \(s \in \mathcal{S}\):
        optional internal state (e.g., residual release budget,
        discrete mode, etc.).
        This allows vesicles to implement multi-stage or stateful
        interaction protocols, rather than a single one-shot effect.
\end{itemize}

The space of vesicle states is the product
\begin{equation}
    \mathcal{Z}
    = \mathbb{R}^{d_c}
      \times \{1,\dots,K\}
      \times V
      \times \mathbb{R}^+
      \times \mathcal{S}.
\end{equation}
In particular, \(\mathcal{Z}\) is independent of the dimension of
activations or parameters in the base network; this decoupling is
crucial for viewing vesicles as first-class entities rather than
hidden tensor components.

\subsection{Configuration of Vesicles}

At discrete time \(t \in \mathbb{N}\) (e.g., within or across training
steps), the system maintains a finite multiset of vesicles
\begin{equation}
    \mathcal{V}_t
    = \big\{v_{t}^{(1)}, v_{t}^{(2)}, \dots, v_{t}^{(N_t)}\big\},
\end{equation}
where \(N_t\) is the number of active vesicles at time \(t\).
The cardinality \(N_t\) can vary over time: in some regimes the system
may maintain only a handful of vesicles; in others, it may deploy a
large ``cloud'' of vesicles that densely explore the network.

We denote the joint state of network and vesicles as
\begin{equation}
    S_t = \big(\theta_t, \{h_t^{(l)}\}_{l=0}^L, \mathcal{V}_t \big).
\end{equation}
Here \(\theta_t\) and \(h_t^{(l)}\) may evolve due to training,
vesicle interactions, or external updates.
Conceptually, Neuro-Vesicles turn the pair \((\theta_t, \{h_t^{(l)}\})\)
into a \emph{medium} on which a second, richer dynamical system
\(\mathcal{V}_t\) lives and acts.

\section{Emission, Migration, Docking, Release, and Decay}

We now specify the vesicle dynamics as a composition of stochastic
kernels acting on the configuration \(\mathcal{V}_t\).
The main design principle is to mirror the qualitative stages of
biological neuromodulation (release, diffusion, binding, action, and
clearance), while retaining differentiability and compatibility with
standard training pipelines.

\subsection{Emission Kernel}

Intuitively, emission answers the question:
\emph{when and where should a vesicle be created, and with what
content and type?}
In biological systems, neuromodulators are often released in response to
activity patterns, reward signals, or global brain states.
We encode this idea via a learned emission intensity.

For each node \(\ell \in V\) and type \(\kappa\), we define an emission
intensity \(\lambda_{\text{emit}}(\ell,\kappa,t)\), which may depend
on local activations, gradients, or meta-variables.
A simple parametrization is
\begin{equation}
    \lambda_{\text{emit}}(\ell,\kappa,t)
    = \sigma\!\Big(
        u_{\kappa}^\top
        \psi_{\text{emit}}\big(h_t^{(\ell)}, g_t^{(\ell)}, m_t\big)
      \Big),
    \label{eq:lambda_emit}
\end{equation}
where
\begin{itemize}[leftmargin=2em]
    \item \(u_{\kappa} \in \mathbb{R}^{d_e}\) is a type-specific vector,
    \item \(\psi_{\text{emit}}\) is an encoder producing features of
          local activations \(h_t^{(\ell)}\), local gradients
          \(g_t^{(\ell)} = \nabla_{\theta^{(\ell)}} L_t\), and meta
          state \(m_t\),
    \item \(\sigma(z) = 1/(1+e^{-z})\) is a sigmoid.
\end{itemize}
Thus, emission intensity is a learned function of both forward and
backward signals, allowing vesicles to be spawned preferentially in
regions that are surprising, high-error, or otherwise ``interesting''
for the task.

Conditioned on \(\lambda_{\text{emit}}(\ell,\kappa,t)\), we can model
the number of new vesicles of type \(\kappa\) emitted at node \(\ell\)
as a Poisson random variable
\begin{equation}
    N^{\text{new}}_{\ell,\kappa}(t)
    \sim \text{Poisson}\big(\lambda_{\text{emit}}(\ell,\kappa,t)\big).
\end{equation}
Other emission count distributions (e.g., Bernoulli or bounded count
models) are possible; we adopt a Poisson form to emphasize the
connection with point processes.

For each such new vesicle, we sample its content, lifetime, and
internal state from emission distributions:
\begin{align}
    \mathbf{c}_0
    &\sim P_{\mathbf{c}}\big(\cdot \mid \ell,\kappa,t\big),
    \\
    \tau_0
    &\sim P_{\tau}\big(\cdot \mid \ell,\kappa,t\big),
    \\
    s_0
    &\sim P_{s}\big(\cdot \mid \ell,\kappa,t\big).
\end{align}
These distributions can themselves be parameterized neural networks,
linking vesicle content to the context that triggered emission.

The emission kernel is then
\begin{equation}
    P_{\text{emit}}(\mathcal{V}_{t}^{+} \mid \mathcal{V}_t, S_t),
\end{equation}
where \(\mathcal{V}_{t}^{+}\) denotes the union of existing and newly
emitted vesicles.
From a systems viewpoint, this turns each node into a potential
generator of information-carrying entities that will later roam the
network and intervene in its computations.

\subsection{Migration Kernel}

After emission, vesicles must move.
Migration controls how vesicles explore the network topology and which
regions they are likely to reach, thereby shaping the spatial pattern
of neuromodulatory influence.

Let \(A \in \{0,1\}^{|V| \times |V|}\) be the adjacency matrix of
\(G\), where \(A_{ij} = 1\) if there is an edge from node \(i\) to
node \(j\).
For each type \(\kappa\), we define a (possibly learnable) transition
matrix \(T^{(\kappa)} \in [0,1]^{|V|\times |V|}\) such that
\begin{equation}
    T^{(\kappa)}_{ij}
    = 0 \quad \text{if } A_{ij} = 0,
    \qquad
    \sum_{j} T^{(\kappa)}_{ij} = 1.
\end{equation}
This captures the ``default'' migration policy for vesicles of type
\(\kappa\), constrained by the network topology.

Given a vesicle \(v_t = (\mathbf{c}_t, \kappa, \ell_t, \tau_t, s_t)\),
its new location \(\ell_{t+1}\) is sampled as
\begin{equation}
    \ell_{t+1} \sim
    P_{\text{move}}\big(\cdot \mid \ell_t,\mathbf{c}_t,\kappa,S_t\big),
\end{equation}
where a simple instantiation is
\begin{equation}
    P_{\text{move}}\big(\ell' \mid \ell_t,\mathbf{c}_t,\kappa,S_t\big)
    \propto
    T^{(\kappa)}_{\ell_t,\ell'}
    \cdot
    \exp\Big(
        \gamma_{\kappa}
        \cdot q_{\text{move}}\big(\ell',\mathbf{c}_t,S_t\big)
    \Big),
    \label{eq:move_kernel}
\end{equation}
with \(\gamma_{\kappa}\) a temperature parameter and
\(q_{\text{move}}\) a score function (e.g., depending on local
gradient norm or uncertainty at \(\ell'\)).
In this way, migration can be biased towards regions where intervention
is likely to matter more, while still being rooted in the network's
graph structure.

\subsection{Docking Kernel}

Migration alone does not guarantee interaction.
Docking decides whether a passing vesicle actually engages with a node
or simply passes by.
This is a crucial difference from tensor-based modulation: vesicle
effects are \emph{event-based}, not continuous.

Given a vesicle at location \(\ell_{t+1}\), we define the probability
that it docks at time \(t+1\) as
\begin{equation}
    p_{\text{dock}}
    = P_{\text{dock}}(\text{dock} \mid \ell_{t+1},\mathbf{c}_t,\kappa,S_t)
    = \sigma\!\Big(
        w_{\kappa}^\top
        \psi_{\text{dock}}\big(h_{t+1}^{(\ell_{t+1})}, \mathbf{c}_t, s_t\big)
      \Big),
    \label{eq:dock_prob}
\end{equation}
with parameters \(w_{\kappa}\) and feature encoder \(\psi_{\text{dock}}\).
Here the docking decision can depend on both local activity and vesicle
content, allowing, for example, vesicles that ``recognize'' particular
activity patterns to preferentially dock where those patterns occur.

We then sample a Bernoulli variable
\begin{equation}
    d_{t+1} \sim \text{Bernoulli}(p_{\text{dock}}).
\end{equation}
If \(d_{t+1}=1\), a release event is triggered.

\subsection{Release Operators}

When a vesicle docks at node \(\ell\), it releases its content
\(\mathbf{c}\) via a set of local operators acting on (i) activations,
(ii) parameters, (iii) learning rules, and (iv) external memory.
This is the point where Neuro-Vesicles connect back to the familiar
machinery of deep learning: they do not replace activations and
gradients, but modulate them in structured ways.

\paragraph{Activation-level release.}
Let the pre-activation at node \(\ell\) be \(h^{(\ell)}\).
We define an activation modification operator
\begin{equation}
    \mathcal{R}^{\text{act}}_{\ell}:
    \big(h^{(\ell)},\mathbf{c},\kappa,s\big)
    \mapsto \Delta h^{(\ell)}.
\end{equation}
A simple form is FiLM-style modulation:
\begin{align}
    (\bm{\gamma}_{\ell}, \bm{\beta}_{\ell})
    &= W_{\text{act}}^{(\kappa)} \mathbf{c}
       + b_{\text{act}}^{(\kappa)},
       \label{eq:film_params}
    \\
    \Delta h^{(\ell)}
    &= \bm{\gamma}_{\ell} \odot h^{(\ell)} + \bm{\beta}_{\ell},
       \label{eq:film_update}
\end{align}
where \(W_{\text{act}}^{(\kappa)}\) and \(b_{\text{act}}^{(\kappa)}\)
are learnable, and \(\odot\) denotes elementwise product.
The updated activation is
\begin{equation}
    \tilde{h}^{(\ell)}
    = h^{(\ell)} + \Delta h^{(\ell)}.
\end{equation}
This recovers familiar modulation mechanisms but makes them conditional
on discrete, mobile entities that may or may not be present at \(\ell\)
at a given time.

\paragraph{Parameter-level release.}
At the parameter level, we define
\begin{equation}
    \mathcal{R}^{\text{param}}_{\ell} :
    \big(\theta^{(\ell)},\mathbf{c},\kappa,s\big)
    \mapsto \Delta \theta^{(\ell)}.
\end{equation}
For example, a rank-one update:
\begin{align}
    \mathbf{u}_\ell
    &= U^{(\kappa)}_{\ell} \mathbf{c},
    \\
    \mathbf{v}_\ell
    &= V^{(\kappa)}_{\ell} \mathbf{c},
    \\
    \Delta \theta^{(\ell)}
    &= \eta^{(\kappa)}_{\ell} \,
       \mathbf{u}_\ell \mathbf{v}_\ell^\top,
\end{align}
with step size \(\eta^{(\kappa)}_{\ell}\).
The updated parameters are
\begin{equation}
    \tilde{\theta}^{(\ell)}
    = \theta^{(\ell)} + \Delta \theta^{(\ell)}.
\end{equation}
This provides a mechanism for vesicles to induce localized structural
changes in the parameter space, potentially enabling rapid, targeted
adaptation.

\paragraph{Rule-level release.}
We allow vesicles to modify the local learning rule at node \(\ell\).
Let the base gradient be
\begin{equation}
    g^{(\ell)}_t = \nabla_{\theta^{(\ell)}} L_t.
\end{equation}
We define a vesicle-conditioned gradient
\begin{equation}
    \tilde{g}^{(\ell)}_t
    = \mathcal{R}^{\text{rule}}_{\ell}
        \big(g^{(\ell)}_t,\mathbf{c},\kappa,s\big),
\end{equation}
for example
\begin{equation}
    \tilde{g}^{(\ell)}_t
    = \alpha^{(\ell)}(\mathbf{c}) \odot g^{(\ell)}_t
      + \beta^{(\ell)}(\mathbf{c}),
\end{equation}
where \(\alpha^{(\ell)}(\mathbf{c})\) and
\(\beta^{(\ell)}(\mathbf{c})\) are outputs of learnable functions of
\(\mathbf{c}\).
The parameter update becomes
\begin{equation}
    \theta^{(\ell)}_{t+1}
    = \theta^{(\ell)}_{t}
      - \eta^{(\ell)}_t \tilde{g}^{(\ell)}_t,
\end{equation}
with possibly vesicle-dependent learning rate
\(\eta^{(\ell)}_t = \eta^{(\ell)}(\mathbf{c})\).
This explicitly separates the \emph{what} (the gradient signal) from
the \emph{how} (the update rule), allowing vesicles to shape the latter.

\paragraph{External-memory release.}
Each node \(\ell\) additionally maintains an external memory state
\(M_\ell\).
We define a write operator
\begin{equation}
    M_\ell \leftarrow
    \text{Write}_\ell(M_\ell, \mathbf{c}, \kappa, s),
\end{equation}
and a read operator used during forward/backward computation
\begin{equation}
    r_\ell = \text{Read}_\ell(M_\ell),
\end{equation}
which can be injected into the computation at \(\ell\), e.g.,
\begin{equation}
    h^{(\ell)}_{\text{extended}}
    = h^{(\ell)} \oplus r_\ell.
\end{equation}
By design, the triple
\(\big(\theta, \{M_\ell\}, \mathcal{V}_t\big)\)
forms a richer state than parameters alone.
Vesicles can then be seen as agents that write structured traces into
this topological memory, leaving behind \emph{spatially localized}
signatures of past events.

\paragraph{Combined release effect.}
Let \(\mathcal{V}_{t,\ell}^{\text{dock}}\) be the set of vesicles that
dock at node \(\ell\) at time \(t\).
The total activation update at \(\ell\) is
\begin{equation}
    \Delta h^{(\ell)}_{\text{total}}
    = \sum_{v \in \mathcal{V}_{t,\ell}^{\text{dock}}}
      \mathcal{R}^{\text{act}}_{\ell}
      \big(h^{(\ell)},\mathbf{c}_v,\kappa_v,s_v\big),
\end{equation}
and similarly for parameters and memory.
Thus, multiple vesicles can cooperate or interfere at a single node,
leading to rich combinatorial patterns of modulation.

\subsection{Decay Kernel}

Finally, vesicles must not persist indefinitely, or the system would
degenerate into a static, densely populated field.
Decay enforces a finite temporal horizon for vesicle influence.

Vesicle lifetime \(\tau\) decays over time according to
\begin{equation}
    \tau_{t+1}
    = \tau_t - \Delta t + \epsilon_{\text{noise}},
\end{equation}
where \(\epsilon_{\text{noise}}\) can model stochastic fluctuations.
A vesicle is removed when
\begin{equation}
    \tau_{t+1} \le 0
    \quad \text{or} \quad
    \text{Absorb}(v_{t+1}) = 1,
\end{equation}
where \(\text{Absorb}(\cdot)\) is an optional learned or deterministic
absorption condition (e.g., explicit ``clearance'' operations at
specific nodes).
This ensures that vesicle dynamics remain non-trivial and that the
system does not accumulate obsolete entities.

\section{Coupled Dynamics of Network and Vesicles}

At a high level, the joint system forms a stochastic dynamical process
\begin{equation}
    S_{t+1} \sim \mathcal{K}(S_t),
\end{equation}
where the transition kernel \(\mathcal{K}\) factorizes into:

\begin{enumerate}[leftmargin=2em]
    \item \emph{Base forward pass}:
    \begin{equation}
        h_{t}^{(0)} = x_t,\qquad
        h_{t}^{(l)} = \phi^{(l)}(h_{t}^{(l-1)};\theta_t^{(l)}).
    \end{equation}
    This is the standard computation of the underlying network.

    \item \emph{Emission}:
    \begin{equation}
        \mathcal{V}_t
        \xrightarrow{P_{\text{emit}}}
        \mathcal{V}_t^{+}.
    \end{equation}
    New vesicles are spawned based on current activity and gradients.

    \item \emph{Migration}:
    \begin{equation}
        v_t^{(n)} \xrightarrow{P_{\text{move}}} v_{t+1/3}^{(n)}.
    \end{equation}
    Existing vesicles move across the graph according to their type
    and content.

    \item \emph{Docking and release}:
    \begin{equation}
        v_{t+1/3}^{(n)}
        \xrightarrow{P_{\text{dock}},\ \mathcal{R}}
        v_{t+2/3}^{(n)},\quad
        h_t^{(l)},\theta_t^{(l)},M_\ell
        \mapsto
        \tilde{h}_t^{(l)},\tilde{\theta}_t^{(l)},\tilde{M}_\ell.
    \end{equation}
    Docking events trigger local modifications of activations,
    parameters, learning rules, and memory.

    \item \emph{Decay}:
    \begin{equation}
        v_{t+2/3}^{(n)} \xrightarrow{\text{decay}} v_{t+1}^{(n)},
    \end{equation}
    keeping only vesicles with \(\tau_{t+1}^{(n)} > 0\).

    \item \emph{Parameter update} (for training):
    \begin{equation}
        \theta_{t+1}
        = \theta_t - \eta_t
          \nabla_{\theta_t}
          L\big(\tilde{h}_t^{(L)},y_t\big),
    \end{equation}
    where \(\tilde{h}_t^{(L)}\) is the vesicle-modified output.
\end{enumerate}

Thus, vesicle dynamics and network learning are tightly coupled.
The network provides the substrate and gradients that drive emission
and migration; vesicles, in turn, shape the effective computation and
learning rule experienced by the network.

\section{Density Relaxation}

Exact discrete vesicle dynamics can be expensive or non-differentiable.
We therefore consider a continuous relaxation in which vesicles are
represented by densities.
This serves both as a practical approximation and as a conceptual link
between the discrete entity view and more classical continuous fields.

\subsection{Vesicle Density and Mean Content}

For each node \(\ell\) and type \(\kappa\), define a scalar density
\(\rho_{\ell,\kappa}(t)\) and a mean content vector
\(\mathbf{C}_{\ell,\kappa}(t)\).
Intuitively,
\begin{equation}
    \rho_{\ell,\kappa}(t)
    \approx \mathbb{E}\big[N_{\ell,\kappa}(t)\big],
\end{equation}
and
\begin{equation}
    \mathbf{C}_{\ell,\kappa}(t)
    \approx \mathbb{E}\big[\textstyle\sum_{n\in\mathcal{I}_{\ell,\kappa}(t)}
    \mathbf{c}_t^{(n)} \big],
\end{equation}
where \(\mathcal{I}_{\ell,\kappa}(t)\) indexes vesicles of type
\(\kappa\) at node \(\ell\).
In this picture, vesicles are no longer tracked individually; instead,
we track how much ``vesicle mass'' and content of each type resides at
each node.

\subsection{Dynamics of the Density Field}

We write the density update as
\begin{equation}
    \rho_{\ell,\kappa}(t+1)
    = \rho_{\ell,\kappa}(t)
      + E_{\ell,\kappa}(t)
      - D_{\ell,\kappa}(t)
      + M_{\ell,\kappa}(t),
    \label{eq:density_update}
\end{equation}
where:
\begin{itemize}[leftmargin=2em]
    \item \(E_{\ell,\kappa}(t)\) is the emission term,
    \item \(D_{\ell,\kappa}(t)\) is the decay/absorption term,
    \item \(M_{\ell,\kappa}(t)\) is the net migration term.
\end{itemize}

A simple differentiable parameterization is:
\begin{align}
    E_{\ell,\kappa}(t)
    &= \lambda_{\text{emit}}(\ell,\kappa,t),
    \\
    D_{\ell,\kappa}(t)
    &= \delta_{\kappa}\,
       \rho_{\ell,\kappa}(t),
    \\
    M_{\ell,\kappa}(t)
    &= \sum_{i} \rho_{i,\kappa}(t)
       \tilde{T}^{(\kappa)}_{i\to \ell}
       - \rho_{\ell,\kappa}(t)
         \sum_{j} \tilde{T}^{(\kappa)}_{\ell\to j},
\end{align}
where \(\delta_{\kappa} \ge 0\) is a decay rate and
\(\tilde{T}^{(\kappa)}\) is a differentiable transition matrix (e.g.,
softmax over neighbor scores).
This is a discrete-time reaction--diffusion system over the graph \(G\),
parameterized by neural networks.

In vector form for a fixed type \(\kappa\),
\begin{equation}
    \bm{\rho}_\kappa(t+1)
    = \bm{\rho}_\kappa(t)
      + \bm{\lambda}_{\text{emit},\kappa}(t)
      - \delta_{\kappa} \bm{\rho}_\kappa(t)
      + \big(T^{(\kappa)\top} - I\big)\bm{\rho}_\kappa(t),
\end{equation}
where \(\bm{\rho}_\kappa(t) \in \mathbb{R}^{|V|}\).
This formulation is fully differentiable and can be trained jointly
with the base network using standard gradient descent.

\subsection{Expected Release Under the Density Approximation}

Under the density model, the expected activation update at node
\(\ell\) can be approximated as
\begin{equation}
    \mathbb{E}\big[\Delta h^{(\ell)}_{\text{total}}(t)\big]
    \approx
    \sum_{\kappa}
    \rho_{\ell,\kappa}(t)
    \cdot
    \overline{\mathcal{R}}^{\text{act}}_{\ell,\kappa}
    \big(h^{(\ell)}_t,\mathbf{C}_{\ell,\kappa}(t)\big),
\end{equation}
where \(\overline{\mathcal{R}}^{\text{act}}_{\ell,\kappa}\) is the
mean effect of a vesicle of type \(\kappa\) with average content
\(\mathbf{C}_{\ell,\kappa}(t)\).
Analogous expressions can be written for parameter-level and
rule-level effects.

This yields a fully differentiable surrogate system for training.
Once trained, one can either keep this continuous version, or use it
as a proposal distribution from which discrete vesicles are sampled,
bridging between field-like and particle-like descriptions.

\section{Neuro-Vesicles for Spiking Neural Networks and Neuromorphic Hardware}

The preceding sections treated the base network abstractly, without
assuming a particular neuron model.
We now show that Neuro-Vesicles are especially natural in the context
of spiking neural networks (SNNs) and neuromorphic hardware, where
computation is already event-driven and time-resolved
\cite{maass1997spiking,roy2019towards,rathi2023neuromorphic}.

\subsection{Spiking Dynamics and Eligibility Traces}

Consider a recurrent SNN with neurons indexed by \(i\), membrane
potentials \(u_i(t)\), and spike trains
\(
    s_i(t) = \sum_k \delta(t - t_i^k)
\),
where \(t_i^k\) are spike times.
A simple leaky integrate-and-fire model reads
\begin{equation}
    \tau_m \frac{d u_i(t)}{dt}
    = -u_i(t)
      + \sum_j w_{ij} \, (s_j * \kappa)(t)
      + I_i(t),
\end{equation}
with membrane time constant \(\tau_m\), synaptic kernel \(\kappa\), and
external current \(I_i(t)\).
Spikes are generated when \(u_i(t)\) crosses threshold and then reset.

For learning in SNNs, many biologically motivated rules are
\emph{three-factor} schemes of the form
\begin{equation}
    \Delta w_{ij}(t)
    \propto
    e_{ij}(t)\, m(t),
\end{equation}
where \(e_{ij}(t)\) is an eligibility trace computed from pre- and
post-synaptic activity, and \(m(t)\) is a modulatory signal encoding
reward, surprise, or context
\cite{bellec2020eprop,liu2021mdgl,zhang2023naca,mazurek2025threefactor}.
Eligibility traces evolve according to local dynamics, e.g.,
\begin{equation}
    \tau_e \frac{d e_{ij}(t)}{dt}
    = - e_{ij}(t)
      + F\big(s_i(t), s_j(t)\big),
\end{equation}
where \(F\) encodes spike-timing-dependent plasticity (STDP).

In many existing models, the modulatory term \(m(t)\) is either a
scalar global signal (``dopamine level'') or a fixed field over
neurons \cite{liu2021mdgl,zhang2023naca}.
Neuro-Vesicles provide a way to \emph{materialize} this modulatory
signal as a population of discrete entities moving on the SNN graph.

\subsection{Vesicle-Carried Modulatory Fields}

To couple vesicles to SNN plasticity, we associate with each synapse
\(w_{ij}\) a modulatory field \(m_{ij}(t)\) defined as the aggregated
influence of vesicles in the neighborhood of the synapse.
One concrete instantiation is
\begin{equation}
    m_{ij}(t)
    =
    \sum_{\kappa}
    \sum_{v \in \mathcal{V}_t}
    \mathbf{1}[\ell_v \in \mathcal{N}(i,j)]
    \, \alpha_{\kappa}\big(\mathbf{c}_v, s_v\big),
\end{equation}
where \(\mathcal{N}(i,j)\) is a local region of the graph (e.g., the
pre- and post-synaptic neurons and their immediate neighbors),
\(\mathbf{1}[\cdot]\) is an indicator, and
\(\alpha_{\kappa}\) maps vesicle content and state to a scalar
modulation strength.
The synaptic update then becomes
\begin{equation}
    \Delta w_{ij}(t)
    =
    \eta \, e_{ij}(t)\, m_{ij}(t),
\end{equation}
so that only those synapses lying in the current trajectory of
vesicles receive strong updates.

In this picture, vesicles realize a \emph{structured} modulatory field:
rather than emitting a global broadcast signal, the system can route
modulation selectively along particular pathways. 
For example, vesicles generated in a ``reward'' circuit could migrate
along recurrent loops implicated in the current task, biasing credit
assignment towards these loops while leaving unrelated circuits
untouched, in line with ideas from cell-type-specific neuromodulation
\cite{liu2021mdgl}.

\subsection{Event-Driven Vesicle Dynamics}

Spiking simulations, especially on neuromorphic chips, are often
event-driven: neuron state is updated only at spike times.
Neuro-Vesicles can be integrated into this framework by letting their
dynamics advance only in response to discrete events.
For instance, let \(\mathcal{T} = \{t^1, t^2, \dots\}\) be the ordered
set of spike times in the network.
Between events, vesicles simply age:
\begin{equation}
    \tau^{(n)}(t^{k+1})
    = \tau^{(n)}(t^{k}) - (t^{k+1}-t^{k}),
\end{equation}
and migration/emission/docking are computed only at event times
\(t^{k}\), using the current pattern of spikes in a local window.

This naturally aligns vesicle updates with the temporal granularity of
SNN hardware simulators and avoids unnecessary computation at idle
times.
Moreover, vesicles can themselves emit modulatory spikes into
dedicated channels, so that their actions are indistinguishable from
ordinary spike events at the level of hardware primitives.

\subsection{Mapping to Darwin3 and Other Neuromorphic Chips}

Neuromorphic chips such as TrueNorth~\cite{akopyan2015truenorth},
Loihi~\cite{davies2018loihi}, and Darwin3~\cite{ma2024darwin3} have
demonstrated that large-scale SNNs with on-chip learning are feasible
at competitive energy budgets
\cite{roy2019towards,yao2024spike,enuganti2025review}.
Darwin3, in particular, supports up to 2.35\,million spiking neurons
and over 100\,million synapses per chip, and exposes a domain-specific
instruction set architecture (ISA) that allows flexible on-chip
learning rules and custom synapse models \cite{ma2024darwin3}.

At a high level, Neuro-Vesicles can be compiled onto such hardware by
treating vesicles as \emph{soft processes} that are implemented by
auxiliary neuron populations and configurable synapses:
\begin{itemize}[leftmargin=2em]
    \item \textbf{Emission} can be realized by neuron populations that
          monitor local spikes and error proxies, triggering the
          creation of new ``vesicle neurons'' whose state encodes
          \(\mathbf{c}\), \(\kappa\), and \(\tau\).

    \item \textbf{Migration} corresponds to routing spikes from
          vesicle neurons to different parts of the chip over time.
          On Darwin3, this can be expressed using its packet-based
          routing fabric and ISA-level support for programmable spike
          forwarding.

    \item \textbf{Docking and release} are implemented as synaptic
          connections from vesicle neurons onto target neurons and
          synapses, using the chip's programmable learning engine to
          modulate weight updates and local currents
          \cite{davies2018loihi,ma2024darwin3}.

    \item \textbf{Decay} is handled by letting vesicle neurons follow
          leaky dynamics; once their internal potential falls below a
          threshold, they stop emitting spikes and are effectively
          cleared.
\end{itemize}

For example, suppose Darwin3 supports a generic synaptic plasticity
update of the form
\begin{equation}
    \Delta w_{ij}
    =
    A_{\text{pre}} \, \text{STDP}_{\text{pre}}(t)
    + A_{\text{post}} \, \text{STDP}_{\text{post}}(t)
    + A_{\text{mod}} \, \text{STDP}_{\text{mod}}(t),
\end{equation}
where the last term is gated by a modulatory spike train.
A Neuro-Vesicle implementation can set
\(
    A_{\text{mod}} = \alpha_{\kappa}(\mathbf{c})
\)
for vesicles of type \(\kappa\), and let these vesicles emit spikes
into the modulatory channel whenever they dock near synapse \(w_{ij}\).
Thus, the vesicle dynamics define \emph{which} synapses see strong
modulatory input and \emph{when}, while the on-chip learning engine
handles the local pre/post computations.

Because Darwin3's ISA is explicitly designed to support new learning
rules~\cite{ma2024darwin3}, implementing Neuro-Vesicles does not
require changing the chip, only supplying a compiler and runtime that
map vesicle processes onto instruction sequences.
This is conceptually similar to how existing compilers map abstract
learning algorithms like e-prop onto neuromorphic substrates
\cite{bellec2020eprop}.

\subsection{Algorithmic Synergies}

Embedding Neuro-Vesicles in SNNs and neuromorphic hardware yields
several potential synergies with existing algorithmic work:
\begin{itemize}[leftmargin=2em]
    \item \textbf{Credit assignment.}
          Vesicle-mediated three-factor learning offers a unifying
          view of algorithms such as e-prop~\cite{bellec2020eprop},
          MDGL~\cite{liu2021mdgl}, and NACA~\cite{zhang2023naca}:
          in each case, the modulatory factor can be seen as arising
          from a vesicle field with different emission/migration
          rules.

    \item \textbf{Temporal processing.}
          Recent work shows that neuromodulated oscillations improve
          temporal robustness in SNNs~\cite{yan2025oscillations}.
          Vesicles provide a way to generate and route such
          oscillatory modulation, for instance by letting vesicle
          content encode phase and frequency.

    \item \textbf{Continual learning and task gating.}
          Neuromodulation-assisted credit assignment can mitigate
          catastrophic forgetting by selectively tagging synapses
          with task-specific traces~\cite{zhang2023naca}.
          Vesicles extend this idea by allowing task-specific
          modulatory entities to persist and move between subnetworks,
          effectively implementing mobile task gates.
\end{itemize}

In all these cases, the SNN and chip-level primitives remain largely
unchanged; what is new is the \emph{organizational layer} that decides
where modulatory resources flow over time.
We view Neuro-Vesicles as a candidate abstraction for that layer.

\section{Neuro-Vesicles as an RL-Controlled Overlay}

An alternative view treats vesicle dynamics as a policy in a Markov
decision process layered on top of the base network.
This perspective is particularly natural when one wants vesicle
dynamics to explicitly optimize a long-term objective such as sample
efficiency, robustness, or meta-learning performance.

\subsection{State, Actions, and Policy}

Define the combined state at time \(t\) as
\begin{equation}
    s_t
    = \big(
        x_t, y_t,
        \{h_t^{(l)}\}_{l},
        \theta_t,
        \mathcal{V}_t,
        \{M_\ell\}_{\ell}
      \big).
\end{equation}
We define a vesicle control action \(a_t\) as a collection of
operations:
\begin{equation}
    a_t
    = \big(a_t^{\text{emit}},
           a_t^{\text{move}},
           a_t^{\text{dock}},
           a_t^{\text{release}}\big),
\end{equation}
specifying where to emit vesicles, how to move them, when to dock,
and which release operators to apply.
Each component may itself be structured and high-dimensional.

A parameterized policy \(\pi_\phi\) defines
\begin{equation}
    a_t \sim \pi_\phi(a_t \mid s_t).
\end{equation}
The environment dynamics (base network and stochastic transitions) are
captured in a transition kernel \(P(S_{t+1} \mid S_t, a_t)\).
From this viewpoint, vesicle control is a meta-controller that operates
on top of a differentiable world model (the base network and its
learning dynamics).

\subsection{Objective and Policy Gradient}

Let the per-step reward be a function of performance and regularization,
e.g.,
\begin{equation}
    r_t = -\ell\big(f_{\theta_t}(x_t), y_t\big) - \Omega(\mathcal{V}_t),
\end{equation}
where \(\Omega(\mathcal{V}_t)\) penalizes excessive vesicle usage.
The objective is the expected discounted return
\begin{equation}
    J(\phi)
    = \mathbb{E}_{\pi_\phi}
      \Big[ \sum_{t=0}^{\infty} \gamma^t r_t \Big],
\end{equation}
with discount factor \(0 < \gamma \le 1\).
Using REINFORCE, the gradient is
\begin{equation}
    \nabla_\phi J(\phi)
    = \mathbb{E}_{\pi_\phi}
      \Bigg[
        \sum_{t=0}^{\infty}
        \nabla_\phi \log \pi_\phi(a_t \mid s_t)
        \big(R_t - b_t\big)
      \Bigg],
\end{equation}
where \(R_t = \sum_{k=t}^{\infty} \gamma^{k-t} r_k\) and \(b_t\) is a
baseline (e.g., a learned value function).
Thus, vesicle emission, migration, docking, and release can be learned
to optimize downstream performance, not just through induced gradients
but also through explicit exploration in the space of neuromodulatory
strategies.

\section{Minimal Prototype With Explicit Vesicles}

Finally, we summarize a minimal explicit implementation at a single
time scale (e.g., per training step).
The goal here is not to be optimal, but to demonstrate that the
proposed formalism can be instantiated with standard deep learning
tooling.

\subsection{Configuration Update}

We maintain a list
\begin{equation}
    \mathcal{V}_t
    = \big\{(\mathbf{c}^{(n)}_t, \kappa^{(n)}, \ell^{(n)}_t,
             \tau^{(n)}_t, s^{(n)}_t)\big\}_{n=1}^{N_t}.
\end{equation}
For each training step on a mini-batch \((x_t,y_t)\), we perform:

\begin{enumerate}[leftmargin=2em]
    \item Forward pass through the base network, obtaining
    \(\{h_t^{(l)}\}\) and loss \(L_t\).
    Here, no modification is needed to the original architecture.

    \item Emission:
    sample new vesicles using \eqref{eq:lambda_emit} and the emission
    distributions, and add them to \(\mathcal{V}_t\).
    In a simple implementation, one might cap the maximum number of
    vesicles emitted per layer per step.

    \item Migration:
    for each vesicle \(n\), sample new location using
    \eqref{eq:move_kernel}.
    This can be implemented as a categorical sampling over neighbors.

    \item Docking:
    for each vesicle \(n\), compute docking probability via
    \eqref{eq:dock_prob}; if docked, apply activation/parameter/memory
    release operators and optionally update its internal state
    \(s^{(n)}_t\).
    The release operators can be implemented as small auxiliary
    networks that transform \(\mathbf{c}^{(n)}_t\) into modulation
    parameters.

    \item Decay:
    for each vesicle \(n\), update lifetime and remove it if
    \(\tau^{(n)}_t \le 0\).
    In practice, this corresponds to deleting entries from a list or
    masking them out.

    \item Parameter update:
    compute gradients (possibly rule-modulated) and update \(\theta\)
    with an optimizer of choice.
    The presence of vesicles changes the effective loss surface and
    update rule, but requires no modification to the optimizer itself.
\end{enumerate}

Even this minimal prototype exhibits all defining properties of the
Neuro-Vesicle paradigm:
(i) explicit discrete vesicle entities,
(ii) migration on the network topology,
(iii) event-based docking and release,
and (iv) coupling between vesicle dynamics and learning.
More sophisticated implementations can build on this scaffold by adding
multiple time scales, more expressive release operators, or explicit
meta-learning of vesicle controllers.

\section{Discussion: Gaps in Existing Neuromodulation and Originality of Neuro-Vesicles}

We briefly situate Neuro-Vesicles in the broader landscape of
``neuromodulation-inspired'' mechanisms in deep learning, and clarify
what conceptual gap this framework aims to fill.

Most existing approaches that claim neuromodulatory inspiration reduce
the idea to one of the following patterns:
\begin{itemize}[leftmargin=2em]
    \item scalar or vector gates that multiplicatively scale activations
          or features,
    \item FiLM-style affine transformations conditioned on context,
    \item hypernetworks that generate weights as a function of side
          information,
    \item attention mechanisms that reweight contributions from different
          sources.
\end{itemize}
While powerful and widely useful, all of these mechanisms are ultimately
realized as \emph{additional terms inside the tensor computation}:
they live in the same space, at the same time scale, and with the same
update mechanism as ordinary activations and parameters
\cite{perez2018film,ha2016hypernetworks,lillicrap2020backprop}.
From this perspective, neuromodulation is treated as a ``tensor
decoration''---an extra dimension, an extra channel, or an extra
learned function in the forward pass.

The proposed Neuro-Vesicle framework challenges this implicit assumption
by positing that neuromodulation should instead be understood as a
\emph{separate dynamical system} that:
\begin{enumerate}[leftmargin=2em]
    \item has its own state space (the vesicle configuration),
    \item has its own dynamics (emission, migration, docking, decay),
    \item interacts with the base network through localized, event-based
          release operations.
\end{enumerate}

This leads to several distinctive properties:
\begin{itemize}[leftmargin=2em]
    \item \textbf{Entity-based modulation}: modulatory signals are
          carried by discrete entities that can be counted, tracked,
          and analyzed, rather than being diffused across all neurons.

    \item \textbf{Topological locality}: vesicle effects depend on the
          paths they take on the network graph, making neuromodulation
          sensitive to architecture in a way that pure tensor algebra
          is not.

    \item \textbf{Temporal extension and memory}: vesicles have
          lifetimes and internal states, allowing them to implement
          multi-step protocols and leaving traces in external memory.

    \item \textbf{Decoupled learning rules}: by acting at the level of
          gradients and update rules, vesicles can reshape how learning
          happens, not just what is represented.
\end{itemize}

Of course, this added expressivity comes with potential drawbacks and
open questions:
\begin{itemize}[leftmargin=2em]
    \item \textbf{Complexity}: the vesicle state space is large, and naïve
          implementations may incur non-trivial overhead in memory and
          computation, especially on conventional GPUs.

    \item \textbf{Stability}: if vesicle emission and migration are not
          well regulated, the system may become unstable, with too many
          vesicles or highly chaotic dynamics. Density relaxations and
          RL-based controllers are two possible tools for regularizing
          this behavior.

    \item \textbf{Identifiability}: different combinations of emission,
          migration, and release policies may produce similar behavioral
          effects, making interpretation and analysis challenging.

    \item \textbf{Task alignment}: not all tasks may benefit from such
          a rich neuromodulatory layer; identifying regimes where
          Neuro-Vesicles are strictly beneficial is an important
          empirical question, particularly for temporal processing and
          continual learning in SNNs
          \cite{bellec2020eprop,yan2025oscillations,zhang2023naca}.
\end{itemize}

Despite these challenges, we argue that treating neuromodulation as a
dynamical layer of mobile entities---rather than a minor tweak to
tensor operations---opens a qualitatively new design space for
deep learning systems, with potential implications for continual
learning, credit assignment, hierarchical reasoning, and deployment on
large-scale neuromorphic computers such as Darwin3 and its successors
\cite{ma2024darwin3,roy2019towards,rathi2023neuromorphic}.

\section{Conclusion}

We have proposed Neuro-Vesicles as a concrete, mathematically grounded
realization of the idea that neuromodulation should be modeled as a
dynamical system on top of neural networks, not merely as extra
parameters inside them.
By introducing a population of mobile, discrete vesicle entities with
their own state, dynamics, and interaction operators, we obtain a
framework that simultaneously:
(i) recovers many existing conditioning mechanisms as limiting cases;
(ii) suggests new algorithmic possibilities for spiking neural networks
and neuromorphic hardware; and
(iii) affords a richer vocabulary for thinking about how learning rules
can be reconfigured in space and time.

The present work is intentionally theory-heavy and experiment-light.
Our goal is to articulate a sufficiently detailed formalism that future
empirical work can build upon it: small-scale prototypes on conventional
GPUs, efficient event-driven implementations on neuromorphic chips like
Darwin3, and biologically grounded models connecting vesicle dynamics
to recent theories of credit assignment in the brain
\cite{lillicrap2020backprop,liu2021mdgl,zhang2023naca,bellec2020eprop}.
We hope that this perspective will help bridge the gap between
abstract neuromodulatory ideas and concrete, programmable mechanisms
in artificial neural systems.

\clearpage  

\appendix
\section*{Appendix}
\section{Relation to Nested Learning and Other Modulatory Frameworks}
\label{app:nested}

This appendix clarifies how the proposed Neuro-Vesicle (NV) framework
relates to existing work on nested learning and neuromodulation-inspired
mechanisms in deep learning and neuromorphic computing. In short,
Neuro-Vesicles are \emph{complementary} rather than competing:
nested-learning methods primarily reorganize \emph{optimization} across
multiple time scales, whereas Neuro-Vesicles enlarge the \emph{state
space} of the model with an explicit population of mobile entities that
implement event-based neuromodulation on the network graph.

\subsection{Conceptual Comparison}

Nested learning views a learning system as a hierarchy of nested
optimization problems operating at different time scales: a fast learner
adapts parameters on short horizons, while slower learners adapt the
fast learner, its objectives, or its inductive biases over longer
horizons. In this picture, all relevant quantities---parameters,
optimizer states, meta-parameters---live in a continuous parameter
space and are updated by gradient-based (or gradient-like) methods at
their respective frequencies. A typical nested-learning stack might
consist of: (i) task-level parameters updated every batch, (ii) a
meta-learner updating learning rates, regularization coefficients, or
initializations over many tasks, and (iii) an even slower outer loop
that tunes the meta-learner itself. All of these levels are described
within the same calculus of differentiable optimization: the system
remains a collection of tensors plus associated update rules, with
nesting expressed through compositionality of objectives and gradients.

From this perspective, ``structure'' enters primarily through the way
objectives are factorized and the way gradients are propagated across
levels. There is no separate ontology beyond parameters and their
updates: a meta-parameter is still just a vector or matrix, and its
effect on the fast learner is mediated by differentiable computation
graphs. Even when nested learning is applied to recurrent networks or
spiking networks, the nested components modify how the existing units
learn, not what kinds of entities exist in the state space.

By contrast, Neuro-Vesicles introduce an additional dynamical layer
whose state space is the configuration of vesicles
\(\mathcal{V}_t = \{(\mathbf{c},\kappa,\ell,\tau,s)\}\).
This layer is not a new level of optimization, but a population of
discrete entities that migrate on the network graph and interact with
computation only through \emph{event-based docking and release}. In
other words, NVs change \emph{what} the system is made of, not only
\emph{how fast} different parts of it are optimized. Each vesicle
persists over multiple forward and backward passes, moves between
layers, and decides stochastically whether to intervene at a given
location. The relevant operations are therefore not just gradient steps
but also birth, movement, interaction, and death of individual
particles.

Crucially, the base optimizer may still be standard backpropagation or
any nested learning algorithm; the vesicle dynamics are orthogonal and
can sit \emph{on top of} those choices. A nested learner can update
\(\theta_t\) and meta-parameters that govern \(P_{\text{emit}}\) or
\(P_{\text{move}}\), while the vesicle population \(\mathcal{V}_t\)
evolves as a stochastic process coupled to the network state. This
yields a clear separation of concerns: nested learning organizes the
hierarchy of optimization problems, whereas Neuro-Vesicles enlarge the
state space with explicit mobile entities that realize neuromodulation
as a spatiotemporal process on the graph \(G=(V,E)\). In the limit where
vesicle counts are very high and lifetimes are very short, the NV layer
degenerates to an effectively continuous field that can be absorbed into
tensor-level modulation; in the opposite limit of sparse, long-lived
vesicles, the system resembles a network of neural modules coupled to a
small set of mobile ``agents'' that carry and deploy modulatory
information at discrete times and places---a regime that has no direct
analogue in classical nested-learning formulations.

Seen from this perspective, nested learning sculpts a vector field over
the space of parameters and meta-parameters, while Neuro-Vesicles sculpt
a distribution over \emph{trajectories} of vesicles moving on the
network graph. The two mechanisms therefore operate on different
geometric objects: gradients drive continuous flows in \(\theta_t\),
whereas vesicle birth, migration, docking, and decay induce
piecewise-stochastic jumps in \(\mathcal{V}_t\). When combined, the
overall system lives in a hybrid space \((\theta_t, \mathcal{V}_t)\)
where slow, smooth optimization and fast, event-based neuromodulation
coexist. This hybrid view makes it explicit that one can meta-learn the
statistics of vesicle dynamics (e.g., sparsity, coverage, or preferred
paths on \(G\)) using nested-learning objectives, without ever reducing
vesicles themselves to yet another tensor in the forward pass.

\clearpage 

\subsection{Axes of Difference}

Table~\ref{tab:nested_vs_nv} summarizes high-level conceptual
differences between Neuro-Vesicles, nested learning, and more classical
tensor-level neuromodulation (e.g., FiLM, hypernetworks).

\begin{table}[H]
\centering
\small
\setlength{\tabcolsep}{4pt}
\renewcommand{\arraystretch}{1.15}
\begin{tabular}{>{\raggedright\arraybackslash}p{2.7cm}
                >{\raggedright\arraybackslash}p{4.7cm}
                >{\raggedright\arraybackslash}p{4.7cm}}
\toprule
\rowcolor{headpink}
\textbf{Dimension} &
\textbf{Neuro-Vesicles\textsuperscript{$\dagger$}} &
\textbf{Nested Learning / tensor-level modulation\textsuperscript{$\ddagger$}} \\
\midrule

\textsc{Core object} &
Explicit vesicle entities
\(v = (\mathbf{c},\kappa,\ell,\tau,s)\) forming a
population \(\mathcal{V}_t\). &
Parameters, optimizer states, and meta-parameters arranged
into nested learners; no new first-class entities beyond
continuous parameter vectors. \\[0.35em]

\textsc{View of neuromodulation} &
Event-driven interventions triggered by vesicle docking and
release; modulation is a stochastic dynamical process on
the network graph. &
Deterministic functions inside the forward/backward pass
(e.g., FiLM, hypernetworks, attention) or slower meta-updates
to parameters/optimizers. \\[0.35em]

\textsc{Representation type} &
\emph{Particle-based}: vesicles live on top of the network
graph, can be counted, tracked, and individually inspected. &
\emph{Field-based}: all information is encoded in continuous
tensors and optimizer states. \\[0.35em]

\textsc{Time scales} &
Multiple time scales arise from vesicle lifetimes \(\tau\),
emission intensities \(\lambda_{\text{emit}}\), and migration
dynamics; fast neural computation is shaped by slower vesicle
dynamics. &
Multiple time scales arise from nested optimization levels or
different update frequencies of parameters/meta-parameters. \\[0.35em]

\textsc{Topology usage} &
Migration kernel \(P_{\text{move}}\) and density dynamics
explicitly depend on the graph \(G=(V,E)\), realizing a
reaction--diffusion process over the architecture. &
Topology mainly enters through the static computation graph;
there is no explicit notion of mobile entities moving along edges. \\[0.35em]

\textsc{Interface to learning rules} &
Vesicles can modify activations, parameters, and local learning
rules through \(\mathcal{R}^{\text{act}}_\ell\),
\(\mathcal{R}^{\text{param}}_\ell\),
\(\mathcal{R}^{\text{rule}}_\ell\), and write to external memory
\(M_\ell\). &
Nested learners adjust loss functions, regularizers, optimizers,
or meta-parameters; tensor-level modulation alters activations
via affine transforms, gates, or generated weights. \\[0.35em]

\textsc{Limit cases} &
Dense, short-lived vesicles recover classical tensor-level
conditioning layers; sparse, long-lived vesicles approximate
mobile symbolic agents that edit computation at rare, decisive
events. &
Changing nesting depth alters how many levels of optimization
are modeled, but does not introduce explicit mobile entities
or event-based computation. \\
\bottomrule
\end{tabular}
\caption{High-level comparison between the proposed Neuro-Vesicle
framework and existing approaches based on nested learning or
tensor-level neuromodulation.}
\label{tab:nested_vs_nv}
\vspace{-0.35em}
{\footnotesize
\textsuperscript{$\dagger$}\,NV = Neuro-Vesicles (this work).\;
\textsuperscript{$\ddagger$}\,Includes nested-learning formulations and FiLM/hypernetwork-style tensor modulation.}
\end{table}

\subsection{Mathematical and Dynamical View}

From a mathematical perspective, Neuro-Vesicles and nested learning
emphasize different primitives. Table~\ref{tab:nested_math} contrasts
the two at the level of their formal objects and update rules.

\begin{table}[H]
\centering
\small
\setlength{\tabcolsep}{5pt}
\renewcommand{\arraystretch}{1.12}
\begin{tabular}{>{\raggedright\arraybackslash}p{2.9cm}
                >{\raggedright\arraybackslash}p{4.7cm}
                >{\raggedright\arraybackslash}p{4.7cm}}
\toprule
\rowcolor{headpink}
\textbf{Aspect} &
\textbf{Neuro-Vesicles\textsuperscript{*}} &
\textbf{Nested Learning\textsuperscript{\textsection}} \\
\midrule

State space &
Joint state
\(S_t = (\theta_t, \{h_t^{(l)}\}_{l}, \mathcal{V}_t, \{M_\ell\}_\ell)\)
combining parameters, activations, vesicles, and external memories. &
Hierarchy of parameter and meta-parameter spaces
\(\{\theta^{(k)}_t\}_k\), often organized as nested learners with
different update frequencies. \\[0.35em]

Core dynamics &
Stochastic kernel
\(\mathcal{K}\) factorizing into emission, migration, docking,
release, and decay; vesicles evolve as a Markov process on \(G\). &
Nested optimization problems
\(\min_{\theta^{(K)}} \dots \min_{\theta^{(1)}} L(\theta^{(1)},\dots,\theta^{(K)})\),
with each level updated by an optimizer (gradient-based or not). \\[0.35em]

Coupling to learning &
Local operators
\(\mathcal{R}^{\text{act}}_\ell,
  \mathcal{R}^{\text{param}}_\ell,
  \mathcal{R}^{\text{rule}}_\ell\)
alter activations, parameters, and learning rules in an
event-based way. &
Upper levels influence lower ones through meta-gradients or
implicit differentiation, adjusting objectives, regularizers,
or optimizers over long horizons. \\[0.35em]

Continuous relaxation &
Graph-based reaction--diffusion dynamics for vesicle densities
\(\rho_{\ell,\kappa}(t)\) and mean contents
\(\mathbf{C}_{\ell,\kappa}(t)\), fully differentiable. &
Relaxations typically appear in the form of bi-level optimization
approximations (e.g., truncated backprop, surrogate meta-gradients). \\
\bottomrule
\end{tabular}
\caption{Mathematical and dynamical contrast between Neuro-Vesicles and
nested learning.}
\label{tab:nested_math}
\vspace{-0.35em}
{\footnotesize
\textsuperscript{*}\,Emphasizes an \emph{entity-based} (particle) dynamical view.\;
\textsuperscript{\textsection}\,Emphasizes a \emph{hierarchical} (multi-level) optimization view.}
\end{table}

\subsection{Spiking and Neuromorphic Deployment}

A major motivation for Neuro-Vesicles is their compatibility with
spiking neural networks (SNNs) and neuromorphic hardware. Because
vesicles are defined as mobile, event-driven entities on the network
graph, they map naturally to routed packets or special spike types in
chips such as Darwin3. Table~\ref{tab:nested_hw} focuses on this
deployment angle.

\begin{table}[H]
\centering
\small
\setlength{\tabcolsep}{5pt}
\renewcommand{\arraystretch}{1.1}
\begin{tabular}{>{\raggedright\arraybackslash}p{3.0cm}
                >{\raggedright\arraybackslash}p{4.6cm}
                >{\raggedright\arraybackslash}p{4.6cm}}
\toprule
\rowcolor{headpink}
\textbf{Hardware-related axis} &
\textbf{Neuro-Vesicles on SNN/neuromorphic\textsuperscript{$\star$}} &
\textbf{Nested Learning / standard modulation\textsuperscript{$\circ$}} \\
\midrule

Physical realization &
Vesicles can be implemented as tagged spike packets or messages
carrying payload \(\mathbf{c}\) and type \(\kappa\), routed by the
on-chip network (e.g., Darwin3 ISA). &
Primarily realized in software training loops; hardware mapping
relies on chosen local learning rules (e.g., STDP, three-factor
rules) without explicit packet-level entities. \\[0.35em]

Locality and routing &
Migration kernel naturally matches event routing fabric; vesicles
can follow specific paths, enabling structured topological memory
and localized neuromodulation. &
Topology mainly constrains synaptic connectivity; no explicit
mechanism for mobile ``modulatory packets'' that traverse the chip. \\[0.35em]

Programmability &
Release operators can be bound to microcode or instructions that
modify synaptic weights, neuron parameters, or on-chip learning
rates when vesicles dock. &
Programmability typically lives at the level of which plasticity
rule is implemented in hardware; fine-grained, stateful modulation
must be encoded indirectly in those rules. \\[0.35em]

Use cases &
On-chip continual learning, task-specific modulation, targeted
credit assignment, and structural plasticity controlled by vesicle
trajectories. &
Meta-learning of learning rates, regularizers, or objectives in
offline training; on-chip deployment mainly uses fixed rules
optimized during training. \\
\bottomrule
\end{tabular}
\caption{Comparison of how Neuro-Vesicles and nested learning relate to
spiking neural networks and neuromorphic hardware deployments.}
\label{tab:nested_hw}
\vspace{-0.35em}
{\footnotesize
\textsuperscript{$\star$}\,Entity-based, packet-level neuromodulation
well aligned with event-driven hardware.\;
\textsuperscript{$\circ$}\,Optimization-centric view with less direct
exposure at the packet/spike level.}
\end{table}

\subsection{Orthogonality and Potential Synergy}

Because nested learning reorganizes \emph{how} parameters and
meta-parameters are optimized, while Neuro-Vesicles extend \emph{what}
constitutes the state of the system, the two perspectives are
orthogonal. In principle, vesicle emission and migration kernels
\(P_{\text{emit}}\) and \(P_{\text{move}}\) could themselves form a
slower nested level, or nested objectives could be defined over
long-horizon statistics of vesicle configurations
(e.g., sparsity, coverage of the graph, or information flow patterns).
Conversely, vesicles can endow nested-learning systems with a concrete,
topology-aware mechanism for implementing neuromodulatory signals in
spiking neural networks and neuromorphic hardware.

This separation of concerns---optimization hierarchy versus
entity-based dynamical layer---is precisely what enables
Neuro-Vesicles to serve as a drop-in, extensible neuromodulatory
primitive that can coexist with a wide range of existing training
frameworks, including nested learning, meta-learning, and
three-factor plasticity rules.

\end{document}